\documentclass[wcp]{jmlr}

\usepackage{longtable}

\usepackage{booktabs}
\usepackage{multirow}
\usepackage{adjustbox}
\usepackage{nicefrac}
\usepackage{stmaryrd}

\usepackage{pgfplotstable,booktabs}
\usepackage{filecontents}
\usepackage{pgfplots}

\usepackage{xspace}
\newcommand{\mathsymbol}[2]{\newcommand{#1}{\ensuremath{\mathit{#2}}\xspace}}

\newcommand{\setA}{\mathcal{A}}
\newcommand{\setI}{\mathcal{I}}
\newcommand{\reals}{\mathbb{R}}
\newcommand{\expectation}[1]{\mathbb{E}\left[#1\right]}
\newcommand*{\defeq}{\mathrel{\vcenter{\baselineskip0.5ex \lineskiplimit0pt
			\hbox{\footnotesize.}\hbox{\footnotesize.}}}%
	=}
\renewcommand{\vec}[1]{\boldsymbol{#1}}
\newcommand{\prob}[1]{\mathbb{P}\left[#1\right]}
\mathsymbol{\surrogateLoss}{\mathcal{L}}
\newcommand{\tool}{Run2Survive}
\newcommand{\toolExp}{\tool Exp}
\newcommand{\toolParTen}{\tool PAR10}
\newcommand{\toolPolyLog}{\tool Poly/Log}

\usepackage{ifthen}
\newboolean{blind}
\setboolean{blind}{false}

\jmlrvolume{xx}
\jmlryear{2020}

\title[A Decision-theoretic Approach to Algorithm Selection based on Survival Analysis]{\tool{}: A Decision-theoretic Approach to \\ Algorithm Selection based on Survival Analysis}

\ifthenelse{\boolean{blind}}
{\author{\Name{Anonymous Author(s)} \Email{anon@ymous.net}\\
\addr Unknown Institution
}}
{
\author{\Name{Alexander Tornede} \Email{alexander.tornede@upb.de}\\
\Name{Marcel Wever} \Email{marcel.wever@upb.de}\\
\Name{Stefan Werner} \Email{stwerner@mail.upb.de}\\
\addr Heinz Nixdorf Institute and Department of Computer Science, Paderborn University, Germany
\AND
\Name{Felix Mohr} \Email{felix.mohr@unisabana.edu.co}\\
\addr Universidad de La Sabana, Chia, Cundinamarca, Colombia
\AND
\Name{Eyke H{\"u}llermeier} \Email{eyke@upb.de}\\
\addr Heinz Nixdorf Institute and Department of Computer Science, Paderborn University, Germany
}}

\begin{document}

\maketitle

\begin{abstract}
Algorithm selection (AS) deals with the automatic selection of an algorithm from a fixed set of candidate algorithms most suitable for a specific instance of an algorithmic problem class, where ``suitability'' often refers to an algorithm's runtime. 
Due to possibly extremely long runtimes of candidate algorithms, training data for algorithm selection models is usually generated under time constraints in the sense that not all algorithms are run to completion on all instances. Thus, training data usually comprises \textit{censored} information, as the true runtime of algorithms timed out remains unknown.
However, many standard AS approaches are not able to handle such information in a proper way.
On the other side, survival analysis (SA) naturally supports censored data and offers appropriate ways to use such data for learning distributional models of algorithm runtime, as we demonstrate in this work.
We leverage such models as a basis of a sophisticated decision-theoretic approach to algorithm selection, which we dub \tool{}. Moreover, taking advantage of a framework of this kind, we advocate a risk-averse approach to algorithm selection, in which the avoidance of a timeout is given high priority.
In an extensive experimental study with the standard benchmark ASlib, our approach is shown to be highly competitive and in many cases even superior to state-of-the-art AS approaches.
\end{abstract}

\section{Introduction}\label{sec:introduction}

Algorithm selection (AS) denotes the task of recommending algorithms for specific instances of an algorithmic problem class, such as SAT or integer optimization. More specifically, the goal is to recommend the algorithm that is most suitable for the instance at hand, compared to other algorithms being also available.
This idea is largely motivated by the observation of \textit{performance complimentarity} of algorithms, essentially suggesting that no algorithm dominates all others on all instances \citep{automated_algorithm_selection_kerschkeHNT19}. In this regard, suitability is usually assessed in terms of a measure such as the quality of the solution returned, or the algorithm's runtime in the case of constraint satisfaction (CSP) problems. Here, we focus on the latter, i.e., ``being more suitable'' is understood as having a shorter runtime.  

In this context, a commonly encountered yet largely unsolved problem is the one of \emph{censored training data}.
Common AS approaches leverage observed runtime data together with machine learning methods such as regression \citep{xu2007satzilla,sunny_amadiniGM14}, (pairwise-)classification \citep{satzilla11_xu2011hydra}, or clustering \citep{isac_kadiogluMST10}, in order to predict the runtime or the competitiveness of the algorithms on a new problem instance. 
However, the underlying data, consisting of runtimes measured for a specific algorithm on a certain problem instance, are often incomplete and partial: Not all algorithms have necessarily been run on all instances so far, and even if so, the observed runtimes might be \emph{censored} due to a timeout. In the setting of AS, such timeouts are commonly set to save computational resources.
The aforementioned approaches to runtime prediction have no way of directly dealing with censored data points, and instead apply ad-hoc solutions such as discarding them or imputing missing runtimes by default values. Most of these solutions are lacking a theoretical foundation, and, as will be detailed later on, 
some of them can lead to overoptimistic models. 

On the other side, a well-established class of statistical methods summarized under the notion of \emph{survival analysis} \citep{survial_analysis_kleinbaum2010} is tailored to exactly such kind of censored data. Survival analysis deals with time-to-event data and seeks to model the underlying time-dependent phenomena. 
It originates from applications in medicine, where one is interested in the survival time (or time until death) of patients under various conditions.
Luckily, it is not uncommon that a medical study ends before all patients die.
Hence, the variable of interest, i.e., the survival time, can be censored in the sense of not being precisely observed.
Methods in the field of survival analysis are able to handle such data in a proper way.


In this paper, we leverage survival analysis methods for the purpose of estimating the runtime of algorithms in the context of algorithm selection. More specifically, we make use of runtime data to estimate survival distributions that are specific to combinations of algorithms and problem instances. Since these are \emph{generative} models providing rich information about runtimes in the form of fully specified probability distributions, they can serve as a basis of a sophisticated \emph{decision-theoretic} approach to algorithm selection. Taking advantage of a framework of this kind, we advocate a \emph{risk-averse} approach to algorithm selection, in which the avoidance of a timeout is given high priority. 

\section{The (Per-Instance) Algorithm Selection Problem}\label{sec:algorithm_selection_problem}
In Rice's (per-instance) algorithm selection problem \citep{algorithm_selection_problem_rice76} we are faced with a problem instance space $\setI$ and a set of candidate algorithms $\setA$, and we are interested in finding a mapping $s: \setI \longrightarrow \setA$, called algorithm selector. For a given problem instance $i \in \setI$, the algorithm selector is supposed to choose the algorithm $a^* \in \setA$ optimizing a performance measure $m: \setI \times \setA \longrightarrow \reals$ which is costly to evaluate. Accordingly, the optimal selector, referred to as \textit{oracle} or \textit{virtual best solver} (VBS), is then defined as 
\begin{equation}
    s^*(i) = \arg\min_{a \in \setA} \expectation{m(i,a)}
\end{equation} for all problem instances $i \in \setI$, where the expectation accounts for the fact that $m(i,a)$ is a random variable due to possible randomness in the application of an algorithm.

\subsection{Surrogate Models}

Since $m$ is costly to evaluate and the selection process itself typically underlies a time constraint, one cannot just evaluate $m$ on all algorithms in $\setA$.
Therefore, at the core of the majority of existing AS approaches, machine learning models are employed for learning a surrogate model $\widehat{m}: \setI  \times \setA \longrightarrow \reals$, which is fast to evaluate. Due to this property, an algorithm selector $s: \setI \longrightarrow \setA$ can be specified as 
\begin{equation}\label{eq:selector_based_on_estimate}
    s(i) \defeq \arg\min_{a \in \setA} \widehat{m}(i,a) \, .
\end{equation}
To infer a surrogate model, we are usually provided with a set of instances $\setI_D \subset \setI$ for which the performance in terms of $m$ is known for \textit{some but rarely all} algorithms $a \in \setA$. Furthermore, as a prerequisite for generalizing runtime information across instances, we assume that these can be represented as $d$-dimensional feature vectors using a feature function $f:\setI \longrightarrow \reals^d$. In the remainder of this paper, we write $i$ for an instance and $\vec{i}=f(i)$ for its feature representation.

\subsection{Censored Data}\label{sec:censored_data}
In combinatorial optimization, some algorithms may take extremely long to solve some instances \citep{heavy-tailed_distributions_in_combinatorial_search_gomesSC97}.
Due to this, the generation of training data for surrogate models in algorithm selection is usually time-constrained in the sense that a time limit $C$, called \textit{cutoff}, is set when running an algorithm.
If the algorithm does not solve the instance prior to this timeout, the execution is aborted to save computational costs. In such a case, the runtime data is \textit{right-censored} \citep{survial_analysis_kleinbaum2010} in the sense that $m(i,a) > C$, i.e., the true runtime is known to exceed $C$.
In cases where the data is not explicitly generated for training, but comes from a real-world scenario, it usually features only one true algorithm runtime for each instance, as in the absence of an algorithm selector, it is common practice to run many algorithms in parallel and stop all others as soon as the first one solves the instance.
Hence, the data matrix $M \in \reals^{|\setI_D| \times |\setA|}$ of known performance values spanned by the set of training instances $\setI_D$ and algorithms $\setA$ is only partially filled with known runtimes. 
As an example, consider the algorithm selection benchmark ASlib \citep{aslib_bischlKKLMFHHLT16}, where in some cases more than $70\%$ of the available data points are censored (see Table \ref{tab:aslib_scenarios}).

A na{\"i}ve approach to deal with this problem is to either impute missing runtimes with a default value or ignore them altogether when training a surrogate $\widehat{m}$. Common choices for a default are the cutoff time $C$ or ten times the cutoff time, motivated by the PAR10 score, which is a common evaluation measure in the domain of AS. The PAR10 score corresponds to the runtime if the algorithm did not time out or ten times the cutoff if it timed out. However, all these strategies exhibit considerable drawbacks:
\begin{itemize}
\item
Any form of imputation is a deliberate distortion of the training data and thus should be done with care. In scenarios with many censored data points, e.g., the one with over $70\%$, imputation may lead to strongly biased surrogate models.
More specifically, an imputation of missing values with the cutoff time, which is lower than the actual runtime, will lead to a systematic underestimation of true runtimes. 
\item
Dropping censored samples altogether is a waste of valuable information. Although the censored samples do not inform about precise runtimes, they still carry information, namely that $m(i,a) > C$. Furthermore, by dropping the long and keeping the short runtimes, there is again a danger of inducing overoptimistic models.
\end{itemize}

Although we are not the first to remark these problems of censored data (cf.\ Section \ref{sec:related_work}), very few work has been done on solving them in the context of algorithm selection. A method for imputing censored data points introduced by \cite{regression_with_censored_data_schmee1979simple} was studied in the context of algorithm configuration (AC) \citep{bo_with_censored_data_hutterHL13,benchmarking_algorithm_configurators_eggenspergerLHH18} and algorithm selection \citep{xu2007satzilla}. A generalization of this method proposed by 
\citet{benchmarking_algorithm_configurators_eggenspergerLHH18} starts by fitting a model on the uncensored data points, and then uses it to predict the mean $\mu$ and the variance $\sigma^2$ of the distribution for each censored data point. Based on these statistics, a truncated normal distribution $\mathcal{N}(\mu, \sigma^2)_{\geq C}$, where $C$ is a known lower bound on the true runtime, is computed. Lastly, each censored data point is imputed with the mean of its associated truncated normal distribution and the model is refit based on both the censored and uncensored data points. This process is repeated until a stopping criterion is reached.

Despite the fact that this approach is a common way to deal with censored data in the context of AS and AC, the assumption of a normal distribution for runtimes is arguable and in contradiction to the more common heavy-tail assumption \citep{heavy-tailed_distributions_in_combinatorial_search_gomesSC97}. Moreover, the method was originally introduced for linear models and shown to work well for problems with only a single feature, but there is no strong justification for why the method should work well for higher-dimensional problems and more powerful models, too. In practice, one can observe that the method often fails to improve over training with the censored data directly (see Fig. \ref{fig:censored_data_variants}).

Fortunately, survival analysis offers models which are able to \textit{directly} work with censored data points in their learning procedure, without any need for imputation.

\section{Survival Analysis and Random Survival Forests}\label{subsec:survival_analysis}

In this section, we recall some basic concepts of survival analysis as developed in statistics. Moreover, we present a modern, nonparametric approach to fitting survival distributions, which is based on tree induction and ensemble learning techniques as commonly used in machine learning. 

\subsection{Basic Concepts of Survival Analysis}

In survival analysis (SA), we typically proceed from historical data of the form 
\begin{equation}\label{eq:sa_dataset}
    \mathcal{D} = \left\{\left(\vec{x}_n, y_n,\delta_n\right)\right\}^N_{n=1} \, ,
\end{equation} where $\vec{x}_n \in \mathcal{X} \subseteq \mathbb{R}^d$ is a d-dimensional feature representation of a context,  $y_n \in \mathbb{R}_+$ is the observed time until the event of interest occurred for the given context, and $\delta_n \in \{0,1\}$ indicates whether the sample is (right-)censored ($\delta_n=1$) or uncensored ($\delta_n=0$), i.e., $y_n$ is the true time until the event occurred or a clipped version thereof. More precisely,  
\begin{equation}
    y_n = 
    \begin{cases}
    T_n & \text{ if } \delta_n = 0 \\
    C_n & \text{ if } \delta_n = 1
    \end{cases} \, ,
\end{equation} where $T_n$ is the uncensored survival time and $C_n$ the censored one of instance $n$, which means that $y_n$ is only a latent representation of $T_n$. 
Given such historical recordings, one is interested in inferring a model 
\begin{equation}\label{eq:sa_predictor}
    z: \mathcal{X} \to \reals_+ \, ,
\end{equation} which, given a context $\vec{x}$, correctly predicts the unknown survival time $T \in \reals_+$. 

In our setting of algorithm selection, the event of interest is the termination of an algorithm, and a context is given by a problem instance. Moreover, since we consider a fixed cutoff $C$ for a timeout, $C_n = C$ for all $1 \leq n \leq N$. 

Obviously, the problem of inducing (\ref{eq:sa_predictor}) as stated above is very similar to the problem of standard regression. However, since the target variable is censored, the training data contains partial information and hence uncertainty. Accordingly, to capture this uncertainty, the majority of survival analysis approaches employs probabilistic models. 
To this end, the time until an event of interest $T$ occurs is considered as a random variable, which is modeled via a probability distribution using the so-called \textit{survival function} (SF):
\begin{equation}
    S(t, \vec{x}) = \prob{T \geq t \mid \vec{x}} \, ,
\end{equation} i.e., the probability that the event of interest occurs at time $t$ or later, given context $\vec{x}$. Note that the survival function can be equivalently  expressed in terms of the  cumulative (death) distribution function (CDF) $F(t,\vec{x}) = \prob{T \leq t \mid \vec{x}}$ as $S(t, \vec{x}) = 1 - F(t,\vec{x})$. While the survival function is defined according to the non-occurence of the event until a certain time, the \textit{hazard function} 
\begin{equation}
    h(t, \vec{x}) = \lim\limits_{\Delta_t \to 0} \frac{\prob{t \leq T < t + \Delta_t \mid T \geq t, \vec{x}}}{\Delta_t}
\end{equation}
can be interpreted as expressing a degree of propensity of the event to occur at time $t$, under the condition that it did not occur before. 
The above functions are closely related to each other, and knowing one of them suffices to derive the others:
\begin{equation}\label{eq:survival_hazard_connection}
    S(t, \vec{x}) = \exp\Bigg( -\underbrace{\int_0^t h(u, \vec{x})du}_{H(t, \vec{x})}\Bigg) \, , \quad \quad \quad h(t, \vec{x}) = - \left(\frac{\nicefrac{dS(t, \vec{x})}{dt}}{S(t, \vec{x})}\right)
\end{equation} where $H(t,\vec{x})$ is the \textit{cumulative hazard function}.
To further understand this connection, consider the example of $S(t, \vec{x}) = \exp\left(-\lambda \cdot t\right)$ and $h(t, \vec{x}) \equiv \lambda$, where the hazard rate is modeled as a constant $\lambda$  independent of the context. In practice, this assumption will of course be overly simplistic, and the hazard function will depend on the context features $\vec{x}$ (also called \emph{covariates} in SA). Moreover, to model the dependence on $t$, several parametric families of functions have been proposed, such as log-normal and Weibull functions.

A survival model can be used in various ways to obtain a real-valued prediction of the survival time, as requested by (\ref{eq:sa_predictor}). A natural predictor is the expected survival time, i.e., the expectation of the random variable $T$:
\begin{equation}\label{eq:expectedsurvival}
    z(\vec{x}) \defeq \expectation{T} =  \int\limits_{0}^\infty S(t, \vec{x}) \, dt = \int\limits_{0}^\infty 1 - F(t, \vec{x})\, dt 
\end{equation} 

\subsection{Random Survival Forests}
Due to their excellent predictive power, random forests are often chosen to tackle standard regression problems and also serve as a strong baseline in algorithm selection for modeling a surrogate model $\widehat{m}(i,a)$. For this reason, we chose to model the runtime distribution of an algorithm $a$ via the survival function $S_a$ in the form of a \textit{random survival forest} \citep{ishwaran2008random}, an adaptation of standard random forests for survival analysis. 

While being similar to standard random forests, random survival forests differ in the way they (a) build the individual survival trees and (b) generate predictions from these individual trees. Similar to the CART algorithm \citep{cart_breiman1984classification}, the individual survival trees are binary trees that are built via a recursive splitting approach. Splits are chosen in order to maximize survival difference between child nodes, i.e., by maximizing the difference in observed times $y$ associated with these nodes. This process is continued until at least one of possibly multiple stopping criteria is reached, e.g., a node must contain at least $d_0$ uncensored samples. After such a survival tree has been built, it can be used to estimate the cumulative hazard function $H(t,\vec{x})$ for a query instance $\vec{x}$ in the following way: Starting at the root node, one recursively determines which subtree the query instance $\vec{x}$ belongs to depending on the split criterion of the considered node until a leaf node $e(\vec{x})$ is reached. Let $t_1 < \ldots < t_{N(e(\vec{x}))}$ be the distinct times of events associated with node $e(\vec{x})$, $d_{t,e(\vec{x})}$ be the number of samples in node $e(\vec{x})$ where the event occurred at time $t$, i.e., with $\delta = 0$ and $y = t$, and $Y_{t,e(\vec{x})}$ the number of samples where the event has not occurred until time step $t$, i.e., with $\delta = 1$ or $y > t$. Then, the cumulative hazard function $H(t,\vec{x})$ is estimated using the Nelson-Aalen estimator as follows: 
\begin{equation}
\label{eq:hazardestimate}
    H(t,\vec{x}) = \sum\limits_{t_i < \min\left(t_{N(e(\vec{x}))},t\right)} \frac{d_{t_i, e(\vec{x})}}{Y_{t_i,e(\vec{x})}} 
\end{equation} 
This can be seen as an empirical estimation of the hazard function for each event time $t_i$, and an accumulation of these estimates over time. Note that the cumulative hazard function fully defines the survival distribution for a single survival tree. In order to obtain the distribution based on the entire forest, one simply takes the mean over the cumulative hazard function estimates of the single trees.


\section{Survival Analysis for Algorithm Selection}

To tackle the problem of algorithm selection using SA, we learn one survival distribution, i.e., algorithm runtime distribution, for each algorithm $a$ separately, using the above mentioned random survival forests. To this end, we leverage the training data available by constructing algorithm specific training datasets of the form (\ref{eq:sa_dataset}) as 

\begin{equation}
    \mathcal{D}_a = \big\{ \left( \vec{i}, m(i,a), \llbracket m(i,a) = C\rrbracket \right) \, \vert \, i \in \setI_D \big\} \, ,
\end{equation} where we assume that an occurrence of an algorithm runtime $m(i,a)$ equal to the cutoff time $C$, i.e., $m(i,a)=C$, indicates a timeout of algorithm $a$ on instance $i$ and hence a censored sample. Based on this, we estimate the associated survival distribution $S_a$ for each algorithm $a$, which is fully defined by the associated cumulative hazard function $H_a$.

In practice, the random survival forests estimate survival $\textit{step}$ functions $\widehat{S}_a$ for each algorithm $a$. Accordingly, we can approximate the integral of the survival function as follows: 
\begin{equation}\label{eq:exp_scoring_function_approximation}
    \int_{0}^\infty S_a(t,\vec{i}) \, dt \approx  \sum\limits_{t_k,t_{k+1} \in \{0\} \cup \mathcal{T}_a \cup \{C\}} t_k \cdot \Big[\widehat{S}_a(t_k,\vec{i}) - \widehat{S}_a(t_{k+1},\vec{i}) \Big] \, ,
\end{equation} where $\mathcal{T}_a$ denotes the set of event times, i.e., points of termination, observed for algorithm $a$. In the same way, other integrals related to the survival function, such as the expected runtime, can be approximated.  

\subsection{Decision-theoretic Algorithm Selection}

A survival distribution $S_a$ can be seen as a generative model, which, in contrast to standard regression models, not only provides a point-estimate of an algorithm's runtime, but characterizes the runtime  of algorithm $a$ on problem instance $i$ in terms of a complete probability distribution of a random variable $T_{a,i}$. This is a rich source of information, which can be used to realize algorithm selection in a more sophisticated manner by means of a decision-theoretic approach, adopting the principle of \emph{expected utility maximization} \citep{scho_te82}, or, equivalently, \emph{expected loss minimization}. More specifically, this principle suggests a very natural definition of the algorithm selector $s(\cdot)$ in (\ref{eq:selector_based_on_estimate}):    
\begin{equation}
    \label{eq:lossminimizationproblem}
    s(i) \defeq \arg\min_{a \in \setA} \expectation{ \surrogateLoss(T_{a,i}) }  \, ,
\end{equation}
where $\surrogateLoss: \reals_+ \to \reals_+$ is a \emph{loss function} which maps runtimes to real-valued loss degrees. 

\begin{figure}[t]
    \centering
    \includegraphics[width=0.95\textwidth]{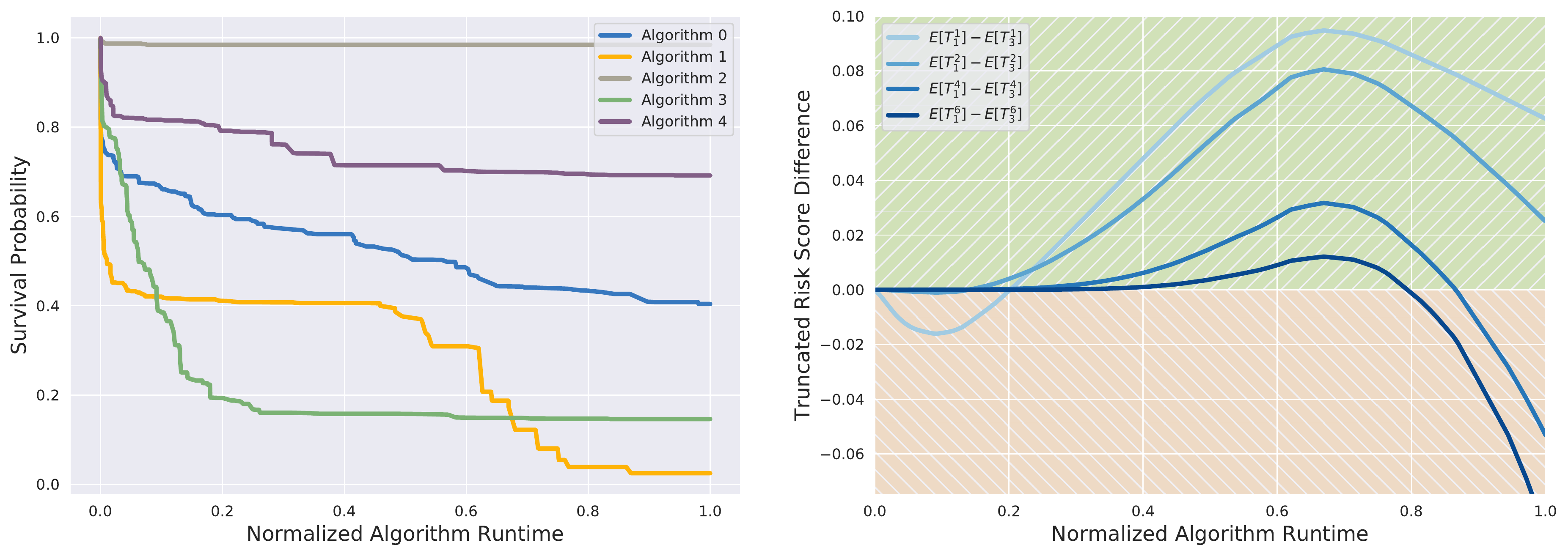}
    \caption{Left: Survival functions of various algorithms on a problem instance. Right: Truncated risk score difference between the runtime of algorithm 1 (yellow on the left) and algorithm 3 (green), i.e. $\expectation{T_1^\alpha \vert T_1 \leq t} - \expectation{T_3^\alpha \vert T_3 \leq t}$. Values below the zero line for the non-truncated score indicate that $\expectation{T_1^\alpha} < \expectation{T_3^\alpha}$ and hence algorithm $1$ is selected over algorithm $3$, which is only the case for larger values of $\alpha$, i.e. with higher risk aversion. Furthermore, it is easy to see that indeed larger values of $\alpha$ emphasize the tail of the distributions as the difference is very close to the zero for small values of $t$, i.e. for the head of the distributions.
    }
    \label{fig:survival_functions_risk_aversion_motivation}
\end{figure}

At first sight, a natural decision criterion is the expected survival time (\ref{eq:expectedsurvival}), i.e., the expected algorithm runtime, which was also suggested by \citet{dynamic_algorithm_portfolios_gaglioloS06} for computing the length of an algorithm schedule. This is obtained as a special case of (\ref{eq:lossminimizationproblem}) with $\surrogateLoss(t) = t$:  
\begin{equation}\label{eq:exp_scoring_function}
     \expectation{\surrogateLoss(T_{a,i})} = \expectation{T_{a,i}} = \int_{0}^\infty S_a(t,\vec{i}) \, dt \, .
\end{equation} 
Upon closer inspection, however, the expected runtime may appear sub-optimal in cases where the performance measure $m$ substantially punishes algorithms running into a timeout. For example, the PAR10 score assigns the runtime of an algorithm as its score if it adheres to the timeout, but 10 times the cutoff if it times out. 
This is not accounted for by the  
expected runtime (\ref{eq:exp_scoring_function}), which considers all regions of the survival function as \emph{equally important} and tends to underestimate the \textit{risk} of a timeout. In fact, an algorithm can have a shorter runtime than another \emph{in expectation}, while having a higher probability of running into a timeout. To see this, consider the comparison between the first and third algorithm in Fig.\ \ref{fig:survival_functions_risk_aversion_motivation} on some example instance; time is normalized to make the cutoff $C=1$. The left plot shows the survival functions of five algorithms computed based on (\ref{eq:survival_hazard_connection}) and (\ref{eq:hazardestimate}). For now, in the right plot, we focus only on the light blue curve, which, for each point in time $t$, depicts how much longer algorithm 1 is \emph{expected} to run compared to algorithm 3, given that both algorithm run for at most $t$, i.e., $\expectation{T_1 \vert T_1 \leq t} - \expectation{T_3 \vert T_3 \leq t}$. Looking at the value of this curve at $t = 1$, it is clear that algorithm 3 has a better expected runtime (\ref{eq:exp_scoring_function}) than algorithm 1. However, the left plot also shows that algorithm 3 has a substantially higher probability to time out, as its probability to terminate at $t=1$, i.e., at the cutoff, is larger than for algorithm 1. While a \textit{risk-averse} view would focus on the survival probability at the timeout and then prefer algorithm 1 over algorithm 3, the expected algorithm runtime prefers 3 over 1, because it weights all parts of the distribution equally\,---\,clearly a sub-optimal choice when timeouts are strongly punished.

As an alternative, the PAR10 score itself may serve as a loss function: 
\begin{equation}\label{eq:par10_surrogate_loss}
    \surrogateLoss(t) = \text{PAR10}(t) = 
    \begin{cases}
    t & \text{ if } t \leq C \\
    10\cdot C & \text{ else }
    \end{cases} \, .
\end{equation} 
The principle (\ref{eq:lossminimizationproblem}) of expected loss minimization then comes down to minimizing the \textit{expected PAR10 score}. If this score plays the role of the \emph{target loss}, on the basis of which the choice of an algorithm will eventually be assessed, this would indeed be a natural idea. It is all the more surprising that, to the best of our knowledge, this approach has not been considered in the literature so far. On the other side, it is true that the PAR10 loss is rather extreme in the sense of strongly focusing on the probability mass close to the cutoff. While we argued that uniformly averaging over the entire range of runtimes (i.e., computing the expected runtime) is sub-optimal, this behavior could be overly conservative and hence not be optimal either. 
To understand this, consider again Fig.\ \ref{fig:survival_functions_risk_aversion_motivation}. The desire to avoid timeouts at any cost would require to choose algorithm 1. At the same time, however, a preference for algorithm 3 appears by no means absurd, since there is undeniably a considerable probability that this algorithm is substantially faster than algorithm 1\,---\,all the more  keeping in mind that the survival functions are only estimates of the underlying ground truth distributions, and hence $\expectation{ \surrogateLoss(T_{a,i})}$ only an estimate of the true expected loss.

\subsection{Risk-Averse Algorithm Selection}

In light of the above discussion, our general goal should be to trade off the different regions of the runtime distributions, keeping in mind the advantages of potentially very short runtimes on the one hand and the risk of timeouts on the other hand. To this end, we consider two parameterized classes of functions for instantiating $\surrogateLoss$. These functions can be seen as \emph{surrogate losses}, i.e., surrogates of the actual target loss (PAR10). Both functions are convex, thereby reflecting a \emph{risk-averse} attitude in decision making \citep{utility_theory_fishburn1969}. 
In the context of algorithm selection, where decisions are algorithms and the avoidance of timeouts has highest priority, risk aversion appears to be a very natural and meaningful property. 
 

The first class of functions consists of polynomials  
\begin{equation}
   \surrogateLoss_\alpha(t) = t^\alpha \, ,
\end{equation}
where $\alpha \in \reals_+$ controls the degree of risk aversion. Hence, larger values of $\alpha \geq 1$ result in stronger risk aversion and are therefore better suited for hard algorithmic problems with a large risk of timeouts, whereas smaller values $0 < \alpha < 1$ should be chosen for simpler problems where the majority of algorithms terminates before the cutoff time. One can observe that the higher $\alpha$, the less important is the behavior of the algorithms for low runtimes (the head of the distribution) and the more important for long runtimes (the tail of the distribution), leading to a clear preference of the ``safer'' alternatives for $\alpha > 2$. In our experimental evaluation, we demonstrate that $\alpha$ can be tuned effectively using hyperparameter optimization techniques.

As a second class of functions, we consider surrogate losses of the form 
\begin{equation}
    \surrogateLoss_{\alpha,\beta}(t) = \min\left[ -\alpha\log(1-t),\beta \right] \, ,
\end{equation} where $\alpha \in [0,1]$ again defines the degree of risk aversion, with smaller values encouraging more risk-averse selections, i.e., potentially safer algorithm choices, and $\beta \in \reals_+$ constitutes an upper bound on $\surrogateLoss$ to limit extreme behavior. Here, we assume that $t \in [0,1]$, which can be achieved via rescaling all runtimes such that $C=1$. Again, both parameters $\alpha$ and $\beta$ can be learned as we demonstrate in the experiments.

In practice, choosing the right surrogate loss and optimally adapting its parameters to the problem at hand is a difficult task. Therefore, we suggest to determine the most suitable surrogate loss in a data-driven way using hyperparameter optimization techniques. In the experimental evaluation, we present an approach based on Bayesian optimization \citep{bo_frazier2018}, automatically selecting either the polynomial or the log-based surrogate loss as well as adequate parameters $\alpha$ (and $\beta$).

\section{Experimental Evaluation}\label{sec:evaluation}
In this section, we provide an extensive evaluation of the proposed methodology based on survival analysis, applying it to the algorithm selection library ASlib v4.0 benchmark suite \citep{aslib_bischlKKLMFHHLT16} and comparing it to state-of-the-art approaches. The benchmark suite comprises a broad variety of scenarios from the literature beyond commonly addressed AS domains.
A comprehensive overview of the considered scenarios is provided in Table \ref{tab:aslib_scenarios}.

\subsection{Experimental Setup}
We evaluate each approach for every scenario conducting a 10-fold cross-validation over the instances of a scenario. Unsolvable instances, for which no valid solution could be determined by any algorithm before violating the fixed timeout, are subsequently removed from the test dataset.

More sophisticated industrial algorithm selection systems commonly integrate complementary techniques such as pre-solvers or feature selectors. For our experimental study, we consider pure algorithm selection models without any pre-solvers or feature selectors in order to focus on the models themselves and their suitability to deal with censored data.

The performance of the approaches is assessed via the \textit{penalized average runtime} (PAR10) metric defined as the average runtime of the selected solver required to solve problem instances. Here, timeout violations are explicitly penalized by a factor of 10. Whether the proposed techniques constitute improvements over the Single Best Solver (SBS) baseline, and consequently close the gap to the Virtual Best Solver, can be assessed in terms of the \textit{normalized PAR10} (nPAR10) metric. Here, the SBS refers to the portfolio's best overall selection, where each algorithm's performance is aggregated over the provided training set's PAR10 scores, whereas the (hypothetical) VBS denotes the optimal algorithm selector.
We define the normalized PAR10 metric as follows:

\begin{table}[t]
    \centering
    \caption{Overview of examined ASlib scenarios including their number of instances (\#I), unsolved instances (\#U), algorithms (\#A) and provided features (\#F). We also show the cutoffs (C) and the percentage of censored algorithm runs (\%C).}
    \label{tab:aslib_scenarios}
    \begin{adjustbox}{width=\textwidth}   
    \begin{tabular}{lllllllllllllllllllllllllll}
\toprule
 & \rotatebox{90}{ASP-POTASSCO}  & \rotatebox{90}{BNSL-2016} & \rotatebox{90}{CPMP-2015} & \rotatebox{90}{CSP-2010} & \rotatebox{90}{CSP-MZN-2013} & \rotatebox{90}{CSP-Minizinc-Time-2016} & \rotatebox{90}{GRAPHS-2015} & \rotatebox{90}{MAXSAT-PMS-2016} & \rotatebox{90}{MAXSAT-WPMS-2016} & \rotatebox{90}{MAXSAT12-PMS} & \rotatebox{90}{MAXSAT15-PMS-INDU} & \rotatebox{90}{MIP-2016} & \rotatebox{90}{PROTEUS-2014} & \rotatebox{90}{QBF-2011} & \rotatebox{90}{QBF-2014} & \rotatebox{90}{QBF-2016} & \rotatebox{90}{SAT03-16\_INDU} & \rotatebox{90}{SAT11-HAND} & \rotatebox{90}{SAT11-INDU} & \rotatebox{90}{SAT11-RAND} & \rotatebox{90}{SAT12-ALL} & \rotatebox{90}{SAT12-HAND} & \rotatebox{90}{SAT12-INDU} & \rotatebox{90}{SAT12-RAND} & \rotatebox{90}{SAT15-INDU} & \rotatebox{90}{TSP-LION2015} \\
\midrule
\multicolumn{1}{l|}{\#I} &         1294 &      1179 &       527 &     2024 &         4642 &                    100 &        5725 &             601 &              630 &          876 &               601 &      218 &         4021 &     1368 &     1254 &      825 &          2000 &        296 &        300 &        600 &      1614 &        767 &       1167 &       1362 &        300 &         3106 \\
\multicolumn{1}{l|}{\#U} &           82 &         0 &         0 &      253 &          944 &                     17 &         117 &              45 &               89 &          129 &                44 &        0 &          456 &      314 &      241 &       55 &           269 &         77 &         47 &        108 &        20 &        229 &        209 &        322 &         17 &            0 \\
\multicolumn{1}{l|}{\#A} &           11 &         8 &         4 &        2 &           11 &                     20 &           7 &              19 &               18 &            6 &                29 &        5 &           22 &        5 &       14 &       24 &            10 &         15 &         18 &          9 &        31 &         31 &         31 &         31 &         28 &            4 \\
\multicolumn{1}{l|}{\#F} &          138 &        86 &        22 &       86 &          155 &                     95 &          35 &              37 &               37 &           37 &                37 &      143 &          198 &       46 &       46 &       46 &           483 &        115 &        115 &        115 &       115 &        115 &        115 &        115 &         54 &          122 \\
\multicolumn{1}{l|}{C}  &          600 &      7200 &      3600 &     5000 &         1800 &                   1200 &       1e+08 &            1800 &             1800 &         2100 &              1800 &     7200 &         3600 &     3600 &      900 &     1800 &          5000 &       5000 &       5000 &       5000 &      1200 &       1200 &       1200 &       1200 &       3600 &         3600 \\
\multicolumn{1}{l|}{\%C} &         20.1 &      28.1 &      27.6 &     19.6 &         70.2 &                     50 &           7.0 &            39.4 &             57.9 &         41.2 &              48.9 &       20 &         60.2 &     54.7 &     55.6 &     36.2 &          24.7 &       60.7 &       33.3 &       47.5 &      53.9 &       66.7 &       50.4 &       73.5 &       23.5 &          9.6 \\
\bottomrule
\end{tabular}

    \end{adjustbox}
\end{table}

\begin{equation}
    \text{normalized PAR10} = \frac{PAR10_{Model} -  PAR10_{VBS}}{PAR10_{SBS} -  PAR10_{VBS}}
\end{equation}
Normalized scores greater than 1 denote that imposed computational costs were on average less efficient than the SBS, whereas scores below 1 indicate an effective utilization of performance complementarity and thus an improvement over the SBS baseline.

Note that the considered approaches require a feature representation of problem instances. Therefore, the time for computing the corresponding features is taken into account as well.

Our experimental study specifically distinguishes the examined baselines according to their performance under different treatment of censored data as discussed in Section \ref{sec:censored_data}. We initially examine which of the previously proposed methods to address censoring complies best with each baseline, and subsequently compare their respectively best results in terms of their PAR10 performance against our proposed methodology. Consequently, each baseline selector is evaluated under removal of censored data (\textit{Ignored}), imputation of the cutoff (\textit{Runtime}), and imputation of timeout penalty (\textit{PAR10}). Where applicable, we also evaluate the method proposed by \cite{regression_with_censored_data_schmee1979simple}. For each baseline selector, the most effective method to cope with censoring is assessed in terms of median PAR10 performance aggregated over all studied ASlib scenarios.

In the overall comparison, we include three approaches leveraging random survival forests for estimating the survival functions. First, we consider an approach selecting algorithms according to their expected runtime as defined in (\ref{eq:exp_scoring_function}) (referred to as \toolExp). Second, a selection according to the expected PAR10 score of an algorithm as defined in  (\ref{eq:par10_surrogate_loss}) (referred to as \toolParTen) is evaluated. Lastly, a method making use of the polynomial and log-based surrogate loss functions is used, where (a) the type of the surrogate loss and (b) respective parameters are tuned by means of Bayesian optimization on the training data (\toolPolyLog).

All experiments were run on machines featuring Intel Xeon E5-2695v4@2.1GHz CPUs with 16 cores and 64GB RAM. 

In order to allow for full reproducibility of our results, the code for all experiments and the generation of plots including detailed documentation can be found on Github\footnote{\ifthenelse{\boolean{blind}}{URL blinded for review}{\url{https://github.com/alexandertornede/algorithm_survival_analysis}}}.

\subsection{Baselines}

In the following, we detail all baselines we compare with.

\textbf{SUNNY} \citep{sunny_amadiniGM14} retrieves the $k$ nearest neighbours of a queried problem instance in terms of Euclidean distance w.r.t.~the feature representation. From this set, the most efficient algorithm in terms of the (possibly imputed) training runtimes is chosen.

\textbf{ISAC} \citep{isac_kadiogluMST10} leverages a g-means clustering algorithm instead of a k-nearest neighbor approach to partition the feature space. Originally designed for algorithm configuration, we transfer ISAC to algorithm selection borrowing SUNNY's decision strategy. More specifically, we first assign the new instance to the closest centroid and subsequently select algorithms performing best on the training instances associated with the respective cluster.

\textbf{SATzilla'11} In contrast, SATzilla'11 \citep{satzilla11_xu2011hydra} decomposes the algorithm selection problem into pairwise comparisons of algorithms. For each pair of algorithms, a Decision Forest is trained to decide which of the two performs better for a problem instance given as input. 
An explicit cost-sensitive loss function weights instances subject to their misclassification costs, i.e., the absolute difference in their imposed computational cost. At prediction time, given a new query instance, the algorithm that wins the highest number of pairwise comparisons is predicted.  

\textbf{PerAlgorithmRegressor} The regression-based PerAlgorithmRegressor baseline learns one Random Forest per candidate algorithm to predict each algorithm's performance while facilitating algorithm selection according to their respective estimates.

\textbf{MultiClassSelector} The MultiClassSelector baseline follows a multi-class classification approach, where instances are first labeled subject to their respectively most effective selections, and a Random Forest is subsequently employed as classifier to establish suitable selections.   

\subsection{Results}

\begin{figure}[t]
    \centering
    \includegraphics[width=0.8\textwidth]{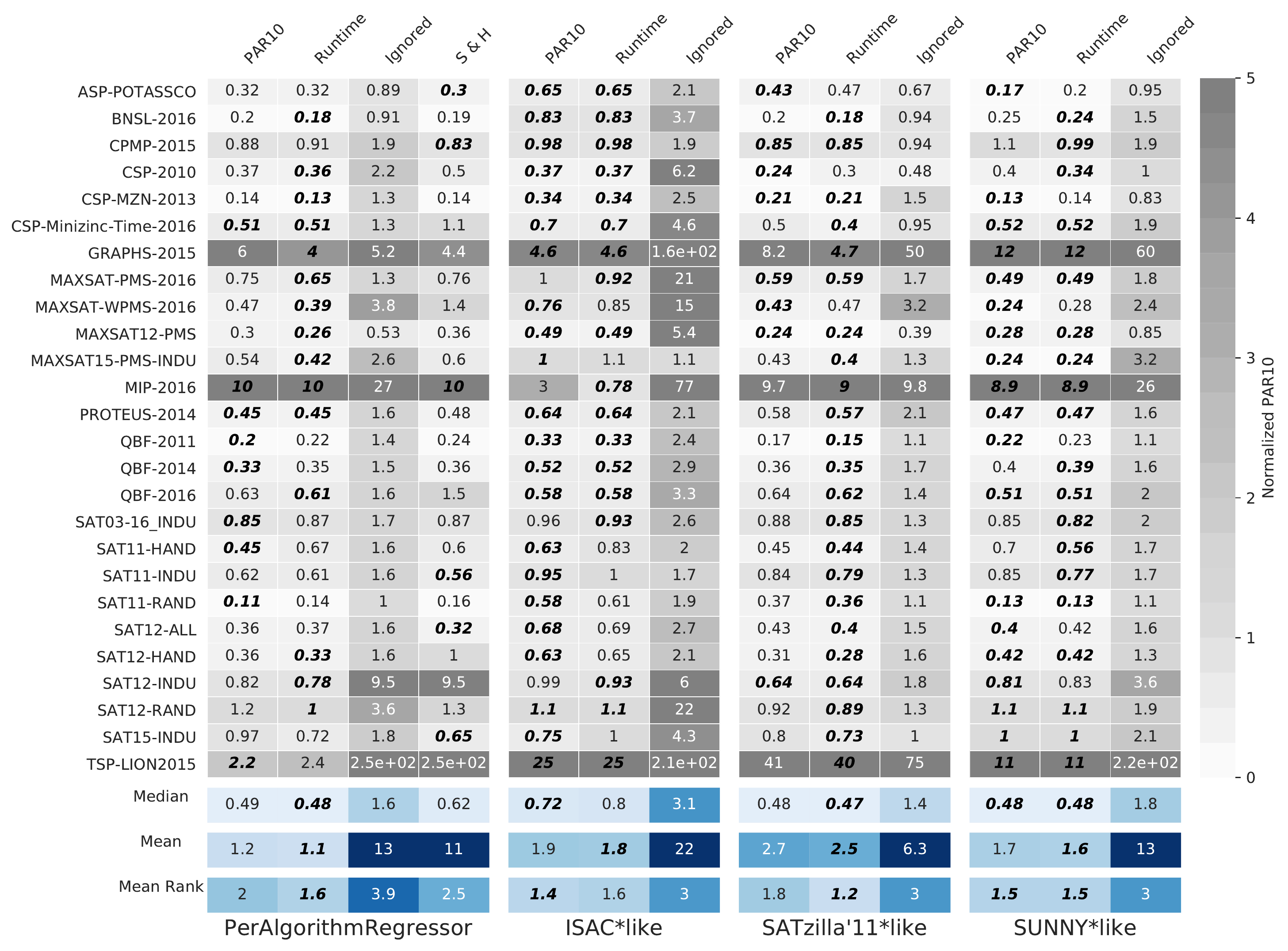}
    \caption{Normalized PAR10 results of baselines for different ways of dealing with censored data: labeling data points as proposed by \cite{regression_with_censored_data_schmee1979simple} (S\&H), with the PAR10 score (PAR10), the cuttoff $C$ (runtime), or the corresponding data points are ignored. The best results for each scenario are printed in bold.}
    \label{fig:censored_data_variants}
\end{figure}

Fig. \ref{fig:censored_data_variants} summarizes the results for the examined baselines subject to different methods of handling censoring. We find that imputing censored algorithm runs with their timeout works best for the PerAlgorithmRegressor, SATzilla'11, and SUNNY, whereas the PAR10 imputation is more effective for ISAC. We further find an iterative adjustment of censored values according to the technique proposed by \cite{regression_with_censored_data_schmee1979simple} to not benefit the PerAlgorithmRegressor in terms of median performance. Since an instance's most efficient solver is preserved regardless of whether censored algorithm runs are imputed with their exact timeout, PAR10 penalty or even entirely ignored, we note that the MultiClassSelector's performance is insensitive to prior imputation methods and thus not included in Fig.\ \ref{fig:censored_data_variants}. While treating censored values as their exact timeout or PAR10 score evidently yields similar results, obviously, discarding censored algorithm runs is the least appealing method, as depending on the ASlib scenario up to 73.5\% of runs are ignored. 

\begin{figure}[t]
    \centering
    \includegraphics[width=0.8\textwidth]{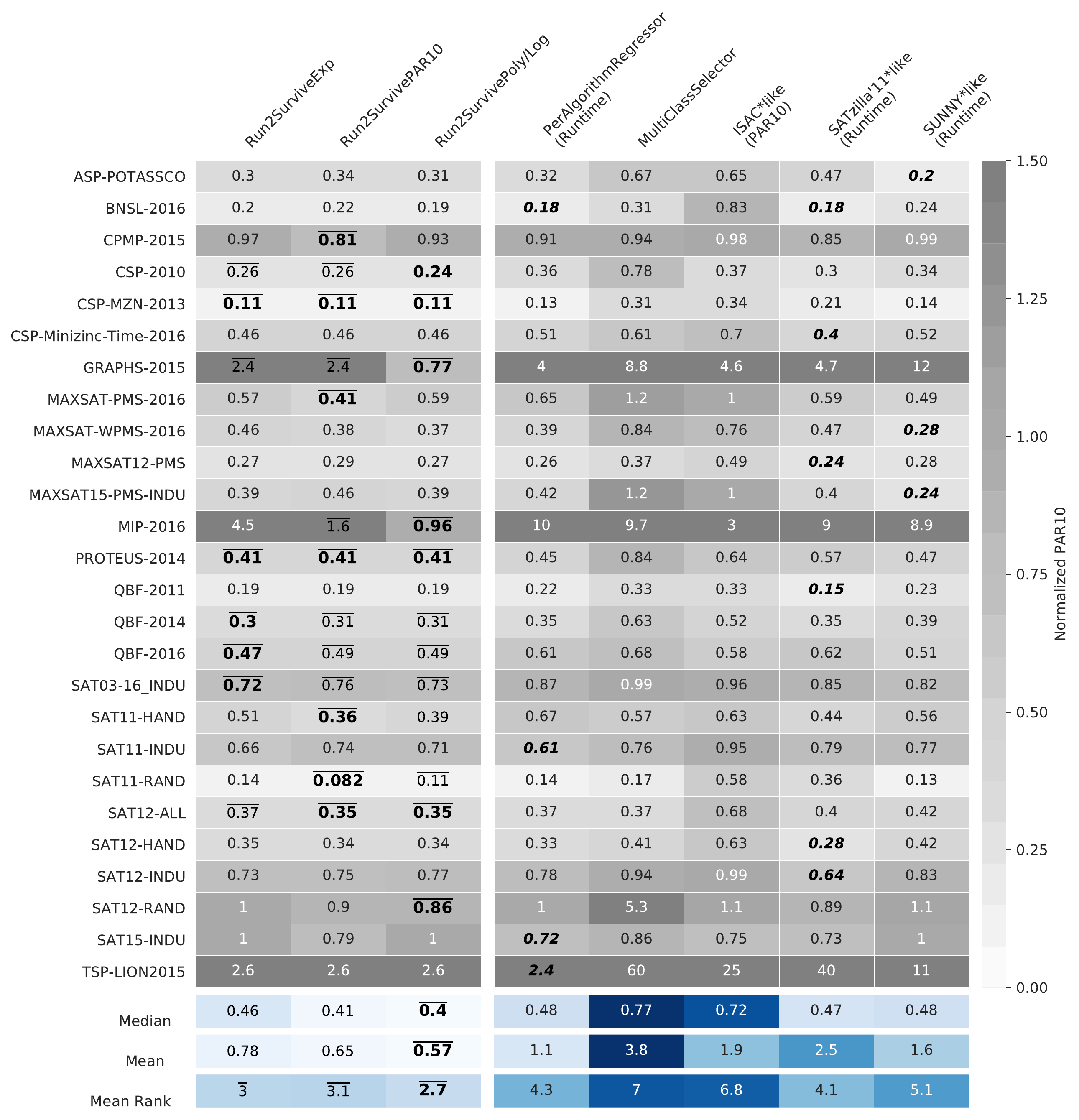}
    \caption{Normalized PAR10 results where for each baseline the way of dealing with censored data is selected according to the minimum median across all examined scenarios. The best results for each scenario are printed in bold whereas an overline indicates beating all baselines.}
    \label{fig:heatmap_main_results}
\end{figure}

Fig.\ \ref{fig:heatmap_main_results} compares the methods based on survival analysis proposed in this paper against the respectively most effective results w.r.t.\ the imputation procedure for each previously discussed baseline. Best results for each scenario are printed in bold, and an overline indicates beating all baselines. Evidently, neither ISAC nor the MultiClassSelector establish competitive algorithm selection on the ASlib benchmark, as the former approach only constitutes an effective choice for one sole scenario, whereas the latter is consistently outperformed. The PerAlgorithmRegressor, SUNNY and SATzilla'11, in contrast, broadly attain competitive results and consequently represent strong competitors to the random survival forests.
However, Fig.\ \ref{fig:heatmap_main_results} illustrates each of our \tool{} variants to outperform their adversaries in terms of median, mean nPAR10, as well as mean rank performance aggregated across all scenarios. Compared to the baselines, \toolExp{} facilitates more effective per-instance algorithm selection on 8 scenarios, whereas \toolParTen{} yields superior results on 13 scenarios. While \toolPolyLog{} beats all baselines in only 12 scenarios, it achieves an nPAR10 score below 1 on all scenarios except one, and hence consistently beats the SBS, which is not the case for any baseline. Furthermore, \toolPolyLog{} beats the SBS in two scenarios where no other approach is able to.

Our findings further illustrate the usefulness of risk-averse decision making in algorithm selection, as \toolExp{} can be improved upon in terms of median and mean nPAR10 performance, and also in mean rank (\toolPolyLog).
In detail, \toolParTen{} and \toolPolyLog{}, respectively, identify superior selection criteria in 10 and 11 ASlib scenarios. However, they also degrade compared to \toolExp{} in 9 respectively 7 cases.

\section{Related Work}\label{sec:related_work}
In this section, we give a brief overview of closely related work. For a more comprehensive overview of algorithm selection and runtime prediction in general, we refer to \citep{automated_algorithm_selection_kerschkeHNT19} and \citep{algorithm_runtime_prediction_hutterXHL14}, respectively.

To the best of our knowledge, the first and only work making direct use of survival analysis in the context of algorithm selection is \citep{dynamic_algorithm_portfolios_gaglioloS06,algorithm_survival_analysis_gaglioloL10}.
\cite{dynamic_algorithm_portfolios_gaglioloS06} present an algorithm called GambleTA computing adaptive algorithm schedules for SAT and the Auction Winner Determination problem instances without any prior learning phase.
GambleTA is a bandit-based algorithm deciding online which algorithms to run for how long such that it can use the obtained data to learn the runtime distributions of the considered algorithms.
These distributions are modeled using cumulative hazard functions (CHF), which in turn are used to compute static schedules based on different techniques, such as selecting at timestep $t$ the algorithm with e.g.\ minimal expected runtime.
Using these static schedules, the authors also show how to construct dynamic schedules.

Our work distinguishes itself from theirs mainly by the following points: Firstly, \cite{dynamic_algorithm_portfolios_gaglioloS06} consider the setting of dynamic algorithm scheduling with a need for online learning, whereas we consider the standard setting of algorithm selection, whence other recommendations are required. Secondly, while the authors only consider two algorithmic problems and algorithm sets of size two, we consider the complete ASlib with over 25 scenarios from different problem domains featuring between 2 and 31 algorithms, making our experimental study much more extensive. Thirdly, the authors consider nearest-neighbor motivated survival analysis models for modeling the CHF of the algorithms, while we make use of random survival forests (although in principle any technique could be used). Finally, and perhaps most importantly, we mainly focus on designing a theoretically sound algorithm selection strategy making use of the concept of risk aversion, leveraging surrogate loss functions in order to better tailor selections towards the structure of loss functions similar to the PAR10 score.

\cite{xu2007satzilla} also remark the problems and implications of censored data in the context of algorithm runtime prediction. Instead of directly applying a method tailored towards such data, they apply the method by \cite{regression_with_censored_data_schmee1979simple} discussed in Section~\ref{sec:censored_data}.
Furthermore, similar to us, they perform a case study on how treating censored data during the training of runtime models impacts these models' performance. They find that (1) both ignoring and imputing censored data with the cutoff time does indeed yield overly optimistic models (as we also note in Section~\ref{sec:censored_data}), and (2) that the method proposed by \cite{regression_with_censored_data_schmee1979simple} yields best performance in terms of the RMSE. 

In the context of algorithm configuration, \citet{bo_with_censored_data_hutterHL13,algorithm_runtime_prediction_hutterXHL14} also note problems arising from censored data when not treating these datapoints appropriately and describe a generalization of the method of \cite{regression_with_censored_data_schmee1979simple} to random forests in order to make these better suited for censored data. Similarly, \cite{parameter_importance_biedenkappLEHFH17} and \cite{benchmarking_algorithm_configurators_eggenspergerLHH18} shortly mention the problem of censored data arising from terminating all running algorithm configurations when one out of many parallel runs finishes, and suggest to impute such censored data by previously discussed means.

\section{Conclusion}

In this paper, we proposed the use of survival analysis techniques combined with a decision-theoretic approach for the problem of algorithm selection.
Taking advantage of the rich information provided by generative models of algorithm runtimes, together with the use of a risk-averse decision strategy to select the most promising algorithm for an unseen instance, we achieved a robust overall performance across different problem domains, such as SAT, CSP, and CPMP.
Applying our method to a suite of 26 benchmark scenarios for algorithm selection from the standard benchmark library ASlib, we find it to be highly competitive and in many cases even superior to state-of-the-art methods.
Moreover, considering several statistics across the considered benchmark datasets, our approach performs best in terms of average rank as well as median/mean normalized PAR10 score across all scenarios, while achieving best performances in up to 13 out of 26 scenarios.

Encouraged by the strong empirical results, we plan to further elaborate on our methodology in future work. Various directions are conceivable in this regard.
First, it would be tempting to transfer it to the extreme algorithm selection (XAS) scenrio \citep{extreme_algorithm_selection_tornedeWH20}, which deals with a significantly larger collection of algorithms to choose from\,---\,correspondingly, XAS requires a feature representation not only for problem instances but also for algorithms.
Another interesting idea is to move to an online setting \citep{online_algorithm_selection_degrooteCBK18}, which calls for the representation of competing risks, since algorithms are stopped based on the earlier termination of their competitors.

\ifthenelse{\boolean{blind}}{}{\acks{This work was partially supported by the German Research Foundation (DFG) within the Collaborative Research Center ``On-The-Fly Computing'' (SFB 901/3 project no.\ 160364472).
The authors also gratefully acknowledge support of this project through computing time provided by the Paderborn Center for Parallel Computing (PC$^2$).}}

\bibliography{main.bib}

\end{document}